\documentclass[letter, 10 pt, conference]{ieeeconf}  

\IEEEoverridecommandlockouts                              

\usepackage{graphics} 
\usepackage{epsfig} 
\usepackage[linesnumbered,ruled]{algorithm2e}
\usepackage{subcaption}
\usepackage{float}
\usepackage{graphicx}
\usepackage{bm}
\usepackage{times}
\usepackage{balance}
\usepackage{amsmath}
\usepackage{breqn}
\usepackage{cite}
\usepackage{threeparttable}
\usepackage{multirow}
\usepackage{booktabs}
\usepackage{bigstrut}
\usepackage{makecell}
\usepackage{color,soul}
\usepackage{blindtext}
\usepackage{amssymb}

\graphicspath{{figures/}}   

\title{\LARGE \bf
    Towards Open-World Human Action Segmentation Using Graph Convolutional Networks
}

\author{Hao Xing$^*$, Kai Zhe Boey$^*$,  Gordon Cheng
    \thanks{Authors are with Institute for Cognitive Systems, School of Computation, Information and Technology, Technical University of Munich, Arcisstraße $21$, $80333$ Munich, Germany. {\tt\small hao.xing@tum.de}, {\tt\small kaizhe.boey@tum.de}, {\tt\small gordon@tum.de}
    }
    \thanks{$*$ Authors contribute equally }
}%

\begin{document}
	
	\maketitle
	\thispagestyle{empty}
	\pagestyle{empty}

    \begin{abstract}
    Current methods for human-object interaction segmentation excel in closed-world settings but struggle to generalize to open-world scenarios where novel actions emerge. Since collecting exhaustive training data for all possible dynamic human activities is impractical, a model capable of detecting and segmenting novel, out-of-distribution (OOD) actions without manual annotation is needed. To address this, we formally define the open-world action segmentation problem and propose a novel framework featuring three key components: 1) an Enhanced Pyramid Graph Convolutional Network with a new decoder for robust spatiotemporal upsampling, 2) hybrid-based training synthesizing OOD data to eliminate reliance on manual labels, and 3) a temporal clustering loss that groups in-distribution actions while distancing OOD samples.
    
    We evaluate our framework on two challenging human-object interaction recognition datasets: Bimanual Actions and Two Hands and Object datasets. Experimental results demonstrate significant improvements over state-of-the-art action segmentation models across multiple open-set evaluation metrics, achieving $16.9\%$ and $34.6\%$ relative gains in open-set segmentation (F1@50) and out-of-distribution detection performances (AUROC), respectively. Additionally, we conduct an in-depth ablation study to assess the impact of each proposed component, identifying the optimal framework configuration for open-world action segmentation.
    
    \end{abstract}
\section{INTRODUCTION}


Human-object interactions (HOIs) play a pivotal role in understanding human activities, providing essential cues for applications such as assistive robotics, healthcare, and autonomous systems. Unlike traditional action recognition, HOI analysis requires identifying human actions and objects while also localizing and understanding their interactions over time, a task known as action segmentation. Besides that, for real-world deployment, especially collaborative systems, they must recognize known interactions while detecting and adapting to novel actions. 

\begin{figure}
\vspace*{0.5\baselineskip}
    \centering
    \includegraphics[width=0.40\textwidth]{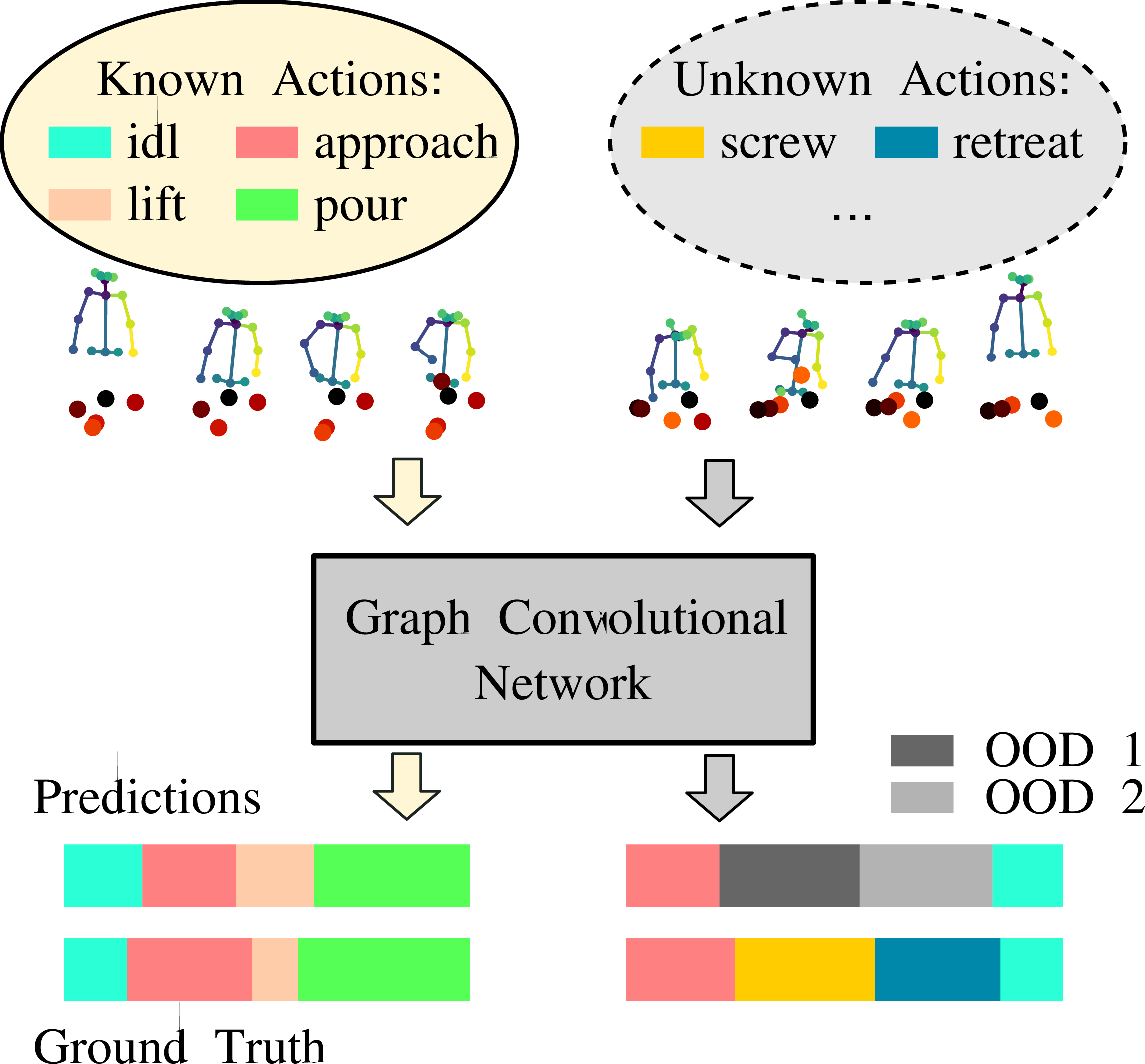}
    \caption{Open-World Human Action Segmentation: detecting and temporally localizing both known and unknown actions.}
    \label{fig:open_world}
\vspace*{-1.5\baselineskip}
\end{figure}

Recently, Graph Convolutional Networks (GCNs) have presented promising results of action segmentation, particularly through skeleton-based representations that offer robustness to occlusions and computational efficiency~\cite{xing2022understanding, Yan2018SpatialTG}. The Pyramid Graph Convolutional Network~\cite{xing2022understanding} improves frame-wise action segmentation through multi-scale feature fusion. However, existing models operate in closed-world settings, where training and testing datasets share the same action categories. This assumption does not hold in real-world scenarios, where models frequently encounter novel actions, leading to poor generalization in open-world settings.

Current approaches to open-world recognition face critical limitations in reflecting true generalization. Existing methods like the Nearest Non-Outlier (NNO) algorithm~\cite{7298799} depend on human-annotated unknown samples during fine-tuning, while ActionCLIP~\cite{Wang2021ActionCLIPAN} leverages CLIP to extend recognition to novel actions, which is a large pretrained visual-language model with inherent semantic knowledge of unknown classes. Although effective, both strategies introduce biases: NNO assumes unrealistic access to labeled unknowns, and ActionCLIP leverages external knowledge from pretrained models. These create an unfair advantage that diverges from real-world open-world constraints. Recent advances, such as the Uncertainty-Quantified Temporal Fusion Graph Convolutional Network~\cite{xing2024understanding}, address Out-of-Distribution (OOD) detection by preserving physical distance in the feature space but overlook inter-class discriminability among OOD actions. 

To address these issues, we redefine open-world action segmentation as a generalization task where models must recognize and segment novel actions using only knowledge from training on closed-world classes, eliminating dependencies on external annotations or pretrained language models, as shown in Fig~\ref{fig:open_world}. We discard the traditional softmax-based classification output for unknown actions classification, which inherently biases predictions toward known classes due to its closed-world confidence calibration. Instead, we explore the relations between feature space and Out-of-Distribution classification, propose a novel Temporal Efficient Upsampling decoder that enables explicit modeling of fine-grained relationships between different actions, and leverages K-means clustering on feature embeddings to categorize OOD actions. In doing so, we propose a Temporal Clustering Loss to enforce tighter feature grouping along the temporal dimension, improving the model’s ability to distinguish In-Distribution from OOD samples. Furthermore, we incorporate Mixup-based augmentation to expose the model to diverse scenarios during training. These innovations ensure more effective OOD detection and classification, enabling robust action segmentation in open-world scenarios.



Overall, the technical contributions of the paper are:

\begin{itemize}
    \item We formally define the problem of open-world action segmentation and establish a structured workflow.
    \item We propose an Enhanced Pyramid Graph Convolutional Network framework with three key components: (i) a novel Temporal Efficient Upsampling decoder for better fusion of multi-scale spatio-temporal features, (ii) a Temporal Clustering Loss that enhances the temporal feature clustering, and (iii) Mixup-based data augmentation to simulate OOD scenarios and improve generalization.
    \item We conduct extensive experiments on two challenging HOI datasets—Bimanual Actions (Bimacs) \cite{Dreher2019LearningOR} and H2O \cite{Kwon_2021_ICCV} and demonstrate that our approach consistently outperforms existing baseline. Additionally, we perform detailed ablation studies to assess the effectiveness of our framework and evaluate alternative design choices.
\end{itemize}

    \section{RELATED WORK}
\label{sec:2}

\subsection{Graph Convolutional Network}

Recently, Graph Convolutional Networks (GCNs) have emerged as a powerful deep learning framework for learning from graph-structured data. Graphs represent non-Euclidean data structures, and conventional neural networks (CNNs) such as Multilayer Perceptrons, Convolutional Neural Networks, or Recurrent Neural Networks are not inherently designed to effectively learn from graph data, as they are primarily tailored for Euclidean data (e.g., images, text, RGB-D videos). There are two main types of GCNs commonly used: 1) Spectral GCNs, which operate in the spectral domain. Bruna et al. \cite{Bruna2013SpectralNA} introduced spectral networks, generalizing CNNs to graphs by leveraging the graph Laplacian spectrum. 
2) Spatial GCNs, which operate directly on the graph domain (nodes and edges). For instance, Hamilton et al. \cite{Hamilton2017InductiveRL} introduced fixed aggregation functions to summarize neighborhood features, while Veličković et al. \cite{velickovic2018graph} proposed Graph Attention Networks (GATs), extending GCNs by incorporating attention mechanisms. 
In the context of skeleton-based action recognition, including our work, the latter approach is predominantly adopted.

Yan et al. \cite{Yan2018SpatialTG} introduced Spatial-Temporal Graph Convolutional Networks (ST-GCN) for skeleton-based action recognition by leveraging spatio-temporal graphs. Shi et al. \cite{2sagcn2019cvpr} extended this with Two-Stream Adaptive Graph Convolutional Networks (2s-AGCN), which adaptively learn joint and bone features. 
More recently, Myung et al. \cite{10478824} introduced Deformable Graph Convolutional Networks, which dynamically learn the most informative joint features in both spatial and temporal domains.

\subsection{Action Segmentation}

Temporal action segmentation (TAS) divides an untrimmed video into segments, each assigned an action label, akin to semantic segmentation but in the temporal domain. It enables automatic action recognition by identifying action onset, progression, and conclusion. Approaches in TAS generally follow three architectural designs: (i) Encoder-decoder architectures, such as the Temporal Convolutional Network (TCN) by Lea et al. \cite{Lea2016TemporalCN}, which leverages temporal convolutions to model long-range dependencies efficiently. (ii) Multistage architectures, exemplified by the Multi-Stage Temporal Convolutional Network (MS-TCN) by Farha et al. \cite{AbuFarha2019MSTCNMT}, which refines predictions iteratively through stacked single-stage TCNs. Filtjens et al. \cite{9998567} extended this with MS-GCN, integrating spatial-temporal graph convolutions for skeleton-based inputs. (iii) Transformer-based architectures, such as ASFormer by Yi et al. \cite{chinayi_ASformer}, which employs self-attention in the encoder and cross-attention in the decoder to capture complex temporal dependencies. This work follows the encoder-decoder architecture for our model design.

\subsection{Human-object interaction (HOI) recognition}

Video-based human-object interaction (HOI) recognition analyzes the temporal structure of untrimmed videos to identify sub-activities and object affordances~\cite{xing2021robust}. Traditional methods, such as Conditional Random Fields, have been largely replaced by deep learning approaches, including CNNs, RNNs, and 3D CNNs, due to their superior ability to model complex relationships. More recently, Graph Convolutional Networks have gained traction for capturing spatial and temporal dependencies in HOI tasks. Morais et al. \cite{9577351} introduced the Asynchronous-Sparse Interaction Graph Network (ASSIGN), which models HOI as a spatio-temporal graph, enabling asynchronous entity updates for improved segmentation but suffering from RNN-related short-term memory limitations. Qian et al. \cite{Qiao2022GeometricFI} proposed the Two-level Geometric feature informed Graph Convolutional Network (2G-GCN), which fuses geometric skeleton-based representations with RGB video features to mitigate occlusion issues. Their fusion-level network integrates an attention mechanism to enhance interaction modeling, leveraging ASSIGN as the backbone for HOI recognition.

\begin{figure*}[t]
  \centering
  \includegraphics[width=0.94\textwidth]{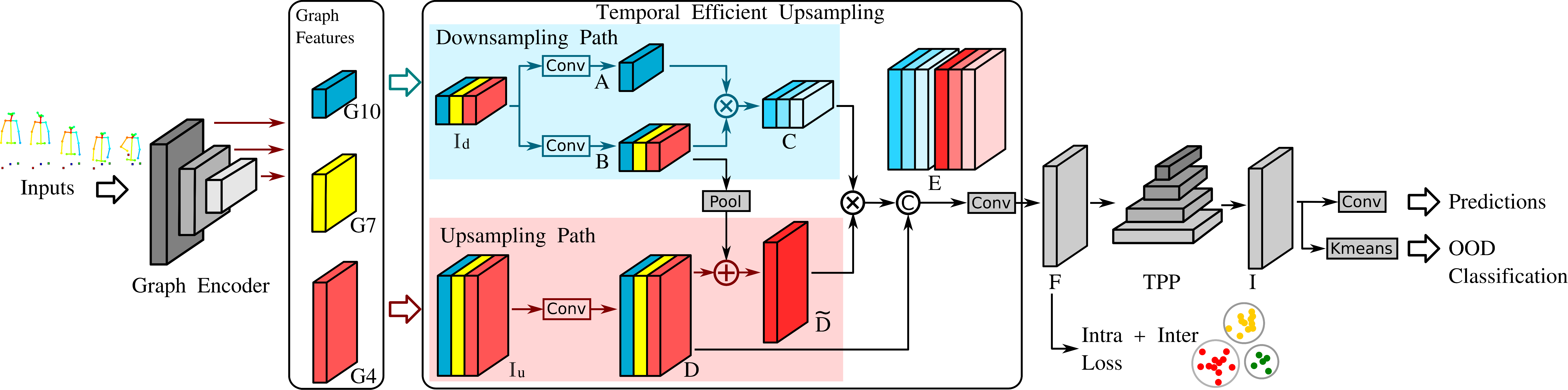}
  \caption{Architecture of the Efficient Pyramid Graph Convolutional Network (EPGCN) with the Temporal Efficient Upsampling (TEU) Module. The TEU decoder processes multi-scale encoder graph features ($G4$, $G7$, $G10$) through dual pathways: (1) the Downsampling Path ($I_d \rightarrow A, B \rightarrow C$); (2) the Upsampling Path ($I_u \rightarrow D \rightarrow \tilde{D}$). The fused output is processed by a Temporal Pyramid Pooling (TPP)~\cite{xing2022understanding} module ($C,\tilde{D},D, \rightarrow E \rightarrow F\rightarrow I$). The produced feature maps $F$ and $I$ are collected for intra/inter-clustering loss optimization and OOD classification, respectively.}
  \label{fig:TEU}
  \vspace*{-1.\baselineskip}
\end{figure*}

\subsection{Generalized Out-Of-Distribution(OOD) Framework}

Out-of-Distribution (OOD) research is critical in deep learning, as models frequently encounter unseen test data that differ from their training distribution, a phenomenon known as distributional shift. This shift is categorized into covariate shift, where ID and OOD samples originate from different domains, and semantic shift, where both share the same domain but belong to different semantic classes. To address these challenges, the generalized OOD framework includes several sub-tasks. (i) OOD detection determines whether a sample belongs to an unknown distribution, with methods such as thresholding softmax probabilities \cite{Hendrycks2016ABF}, while Liang et al. \cite{Liang2017EnhancingTR} improved performance by calibrating softmax outputs and applying temperature scaling. (ii) Open-set recognition extends OOD detection by also classifying known samples. For instance, Bao et al. \cite{BaoICCV2021DEAR} proposed Deep Evidential Action Recognition, leveraging an Evidential Neural Network to predict a Dirichlet distribution over class probabilities. (iii) Generalized zero-shot learning aims to recognize both seen and unseen classes, with approaches such as conditional Wasserstein GANs \cite{Mandal_2019_CVPR} for feature synthesis, although these require prior knowledge of unknown features. (iv) Open-world recognition (OWR), introduced by Bendale et al. \cite{7298799}, requires models to detect and incrementally learn new classes using human annotation by employing the Nearest Non-Outlier algorithm.

Our work follows the OWR paradigm but eliminates the incremental learning stage that relies on human-labeled novel data, which is particularly infeasible for segmentation tasks requiring fine-grained frame-wise labeling. Moreover, we adopt the thresholding approach of Hendrycks et al. \cite{Hendrycks2016ABF}, using output logits to distinguish OOD from ID samples.
    
    \section{Methodologies}
\label{sec:3}

\subsection{Open-World Action Segmentation Workflow}

\textbf{Problem definition and notation}: formally, let $C_{\text{known}}$ denote the set of labeled action classes during training, where all training samples satisfy $ x \in C_{\text{known}}$. In the \textit{Closed-World Action Segmentation (CWAS)} setting, test samples consist only of label-known actions, where all test samples remain within the same distribution as the training set. Conversely, in the \textit{Open-World Action Segmentation (OWAS)} setting, test samples include actions from novel classes $C_{\text{novel}}$ that are disjoint from the known training classes $( C_{\text{novel}} \cap C_{\text{known}} = \emptyset )$. 

The workflow of the Open-World Action Segmentation problem \([x, \phi, F, \nu,\kappa]\) is given by:

\begin{enumerate}
    \item Feature extraction and recognition: let $\phi$ denote a feature extractor that maps an input action sequence $x\in \mathbb{R}^{3\times T\times V}$ to a latent representation $\phi(x)$, where $3, T, V$ denote spatial, temporal size and joints. The recognition function $F$ classifies features into known classes:
    \begin{equation}
        F(x) = arg \max_{c\in C_{known}} f_c(\phi(x)).
    \end{equation}
    
    \item Novelty detection: a detector $\nu$ identifies unknown actions features using a threshold $\alpha$:
    \begin{equation}
        \nu(\phi(x)) = 
            \begin{cases}
              1 \text{ (Known)} & \text{if} \max f_c(\phi(x)) > \alpha\\
              0 \text{ (Novel)} & \text{otherwise}.
            \end{cases}
    \end{equation}
    
    \item Clustering and label assignment:
    For novel samples $\{x | \nu(\phi(x))=0\}$, a clustering function $\kappa$ groups them into distinct pseudo-classes $M$. These are mapped to incrementally indexed new classes $C_{novel} = \{C_{known} + 1, C_{known} + 2, \dots C_{known} + M\}$ via Hungarian algorithm.
    

\end{enumerate}


\subsection{Enhanced Pyramid Graph Convolutional Network}
To realize the OWAS workflow, we introduce an Enhanced Pyramid Graph Convolutional Network (EPGCN). As illustrated in Fig~\ref{fig:TEU}, EPGCN implements an attention based graph convolutional encoder~\cite{xing2022understanding}, introduces a \textit{Temporal Efficient Upsampling} (TEU) decoder combining with \textit{Temporal Pyramid Pooling} (TPP). The encoder-decoder structure form the mapping function $\phi$. The known actions are predicted by a convolutional layer ($F$), while the novel actions are clustered by a K-means algorithm ($\kappa$).

The encoder extracts hierarchical motion representations at three resolutions ($G4$, $G7$, $G10$), corresponding to temporal scales of $T$, $T/2$, $T/4$ for an input sequence of length $T$. The TEU decoder processes these multi-scale features through two parallel pathways: downsampling path and upsampling path. The downsampling path aggregates global context by progressively reducing temporal resolution, enhancing inter-class discriminability for robust novelty detection. The upsampling path recovers fine-grained motion details through learned temporal interpolation, preserving intra-class structural consistency for precise segmentation.

\textbf{Downsampling Path: }The low-level (\(G^4\)) and mid-level (\(G^7\)) features are temporally downsampled via nearest-neighbor interpolation to match the temporal dimension of the high-level feature map (\(G^{10}\)). These are concatenated along the channel dimension:  \(I_{d} = [G_{4}^{\downarrow}, G_{7}^{\downarrow}, G_{10}] \in \mathbb{R}^{N \times C \times T/4 \times V},\) where \(N\) denotes batch size. \(I_{d}\) is refined by two \(1 \times 1\) convolutional layers, generating feature maps \(A\) and \(B\). The feature map \(A\) is normalized using a softmax operation along the temporal dimension for stability:  

\begin{equation}
\tilde{A_{t}} = \frac{\exp{(A_t)}}{\sum_{i=0}^{T/4} \exp({A_i})},
\end{equation}
where \(i\) represents frame indices and \(t\) is a specific frame. The resulting attention map $\tilde{A}$ are applied to $B$ to compute the refined output $C$:  
\begin{equation}
C = \tilde{A}\otimes B,
\end{equation}
with $\otimes$ denoting element-wise multiplication. This emphasizes discriminative temporal regions while suppressing noise.

\textbf{Upsampling Path: }The high-level (\(G^{10}\)) and mid-level (\(G^7\)) features are temporally upsampled to align with the low-level feature map (\(G^4\)). These are concatenated as: \(I_{u} = [G_{4}, G_{7}^{\uparrow}, G_{10}^{\uparrow}] \in \mathbb{R}^{N \times C \times T \times V}\). A \(1 \times 1\) convolution is applied to \(I_{u}\) to produce feature map \(D\). The feature map \(B\) from the downsampling path undergoes temporal average pooling, is replicated across the temporal axis, and fused with \(D\) via element-wise addition:  

\begin{equation}
\tilde{D} = D \oplus \mathcal{P}({B}),
\end{equation}
where $\mathcal{P(\cdot)}$ denotes pooling and replication. A subsequent \(1 \times 1\) convolution enriches \(\tilde{D}\) with high-level semantics, enhancing motion granularity.

\textbf{Multi-Scale Feature Fusion:} The high-resolution feature $\tilde{D}$ and low-resolution context map $C$ (from the downsampling path) are fused via cross-attention, the attention feature map is then channel-wise concatenated with $D$ (the upsampling path’s intermediate representation):


\begin{equation}
E = Concat(C^T \tilde{D}, D).
\end{equation}
This design combines structurally rich low-level features with semantically rich high-level features, enhancing temporal segmentation accuracy. 

The output from TEU is forwarded to a Temporal Pyramid Pooling (TPP)~\cite{xing2022understanding} module to capture global context efficiently. It applies temporal average pooling over hierarchically divided time segments, reducing complexity while preserving key temporal patterns. The final feature maps are collected by the K-means algorithm for OOD classification.

\subsection{Temporal Clustering Loss}
To enforce temporally consistent and discriminative feature clusters for open-world generalization, we propose the Temporal Clustering Loss, a distribution-aware contrastive objective inspired by supervised contrastive learning~\cite{khosla2020supervised}. Unlike conventional contrastive losses that operate on individual samples, our formulation explicitly models class-wise distributions in the spatio-temporal feature space. This is critical for action segmentation, where intra-class temporal variability and inter-class similarity are key challenges.

Let $\mathcal{F}_i = \{f_t\}_{t=1}^T$ denote the temporal sequence of embeddings for class $i \in N$, where $f_t\in \mathbb{R}^d$ is the frame-wise feature at time $t$.

\textbf{Intra-Class Compactness}: compute the dynamic class mean $\bar{\mu}_i$ as an exponentially weighted average of historical and current batch statistics to stabilize training:
\begin{equation}
     \bar{\mu}_i^k = \gamma\bar{\mu}_i^{k-1}+(1-\gamma)\frac{1}{T}\sum_{t=1}^Tf_t,
\end{equation}
where $k$ is the batch index, $\gamma$ controls the momentum. The intra-class loss minimizes the deviation of embeddings from their class mean:
\begin{equation}
    \mathcal{L}_{intra} = \frac{1}{N}\sum_{i\in N}\frac{1}{T}\sum_{f_t\in\mathcal{F}_i}\|f_t-\bar{\mu}_i\|^2.
\end{equation}

\textbf{Inter-Class Separability}: to maximize separation between class distributions, we penalize proximity of pairwise class means:
\begin{equation}
    \mathcal{L}_{inter} = \frac{1}{N}\sum_{i\in N}\sum_{i\neq j}(\|\bar{\mu}_i - \bar{\mu}_j\|^2 + \delta)^{-1},
\end{equation}
where $\delta$ enforces a minimum margin between clusters.

The overall training objective combines classification accuracy with feature clustering constraints:  

\begin{equation}
    \mathcal{L} = \underbrace{\mathcal{L}_{CE} (x,y)}_{\text{Classification}} + \beta\underbrace{ (\mathcal{L}_{\text{intra}} + \mathcal{L}_{\text{inter}})}_{\text{Feature Clustering}},
\end{equation}  
where \(\mathcal{L}_{CE}\) is the cross-entropy loss, ensuring classification accuracy, $\beta$ balances the contributions. By jointly optimizing discriminative classification and geometrically structured embeddings, the model learns temporally stable features that generalize to novel actions while avoiding overconfidence on outliers.


\subsection{Mixup}

We adopt the \textit{Mixup}~\cite{Zhang2017mixupBE} data augmentation technique to enhance model robustness and generalization to Out-of-Distribution (OOD) samples by enforcing linear feature transitions between classes. Unlike traditional empirical risk minimization, which learns only from observed training samples, Mixup trains the model on convex interpolations of input-label pairs, explicitly regularizing the feature space geometry. For spatio-temporal skeleton sequences, this is critical as OOD actions often manifest as semantic interpolations between known classes (e.g., a mix of running and jumping).

Given two randomly sampled skeleton sequences \((x_i, y_i)\) and \((x_j, y_j)\), Mixup generates synthetic training instances:

\begin{equation}
\begin{cases}
\tilde{x} = \lambda x_i + (1 - \lambda) x_j, \\
\tilde{y} = \lambda y_i + (1 - \lambda) y_j,
\end{cases}
\end{equation}

where \(\lambda \sim \text{Beta}(\alpha, \alpha)\) and \(\alpha = 0.2\) controls the interpolation strength. Lower $\alpha$ skews $\lambda$ towards extremes (0 or 1), preserving semantic coherence in skeleton sequences while still expanding vicinal distributions. By leveraging Vicinal Risk Minimization, Mixup expands the learned feature space, mitigating overfitting and improving generalization compared to traditional Empirical Risk Minimization.

\begin{table*}[hbtp]
\begin{center}
  \centering
  \caption{Comparison of open-set and frame-wise performance (F1@K score) of our proposed frameworks against the baseline PGCN and its variations$^1$.}
  \label{table:bimacs}
  \begin{tabular}{l|c|c c c c| c c c}
    \toprule
    \multicolumn{1}{l|}{\multirow{2}{*}{Configurations}} & \multicolumn{1}{c|}{Closed-Set} & \multicolumn{4}{c|}{Open-Set}  & \multicolumn{3}{c}{Out-of-Distribution}  \\
    \multicolumn{1}{c|}{} & \( ACC_1 \) & \( ACC_2 \) & \( F1@10 \) & \( F1@25 \) & \( F1@50 \) & \( AUROC \) & \( ACC_{\text{OOD}} \) & \( h_{\text{score}} \) \\
    \midrule
    PGCN \cite{xing2022understanding} (Baseline) & 79.61 & 70.39 & 83.70 & 81.88 & 73.10 & 50.00 & $\mathbf{85.86}$ & 63.21 \\
    PGCN + Mixup & 79.09 & 71.19 & 90.94 & 89.80 & 83.74 & 52.73 & 67.97 & 63.28 \\
    PGCN + \( \mathcal{L}_{TC} \) & 82.09 & 80.68 & 90.66 & 89.01 & 81.95 & 75.48 & 57.30 & 65.14 \\
    PGCN + Mixup + \( \mathcal{L}_{TC} (4th) \) & 82.86 & 84.25 & 94.34 & 92.39 & 86.26 & $\mathbf{84.66}$ & 63.08 & 72.30 \\
    PGCN + Mixup + \( \mathcal{L}_{TC} \) & $\mathbf{86.08}$ & $\mathbf{86.05}$ & $\mathbf{96.07}$ & $\mathbf{95.31}$ & 89.33 & 84.10 & 73.59 & 78.50 \\
    TEU$^2$ & 76.80 & 68.13 & 92.13 & 90.33 & 80.81 & 50.00 & 73.33 & 60.57 \\
    TEU + Mixup & 82.96 & 74.67 & 90.53 & 89.12 & 81.50 & 52.85 & 63.80 & 64.57 \\
    TEU + \( \mathcal{L}_{TC} \) & 78.15 & 76.70 & 88.81 & 87.43 & 78.47 & 72.33 & 77.43 & 75.12 \\
    TEU + Mixup + \( \mathcal{L}_{TC} \) (\textbf{EPGCN}) & 85.21 & 85.71 & 95.10 & 95.08 & $\mathbf{90.03}$ & 84.62 & 84.69 & $\mathbf{84.65}$  \\
    \bottomrule
  \end{tabular}
    \begin{tablenotes}
        \item[a]$^1$All configurations are evaluated on the Bimacs~\cite{Dreher2019LearningOR} subject 1 testset. The best results across all modifications are highlighted in \textbf{bold}.
        \item[a]$^2$The encoder is from PGCN, and TEU is the decoder.
  \end{tablenotes}
\end{center}
  \vspace*{-1.5\baselineskip}
\end{table*}

    \section{EXPERIMENTS AND RESULTS}
\label{sec:4}

\subsection{Dataset}


The \textbf{Bimanual Actions (Bimacs)} \cite{Dreher2019LearningOR} dataset comprises 540 RGB-D recordings (2 hours and 18 minutes) at 15 FPS, capturing bimanual tasks like pouring milk while stirring cereal in a kitchen and workshop. It includes 26 nodes (12 for human joints and 14 for object centers), and framewise annotations for 12 objects and 6 subjects, covering 14 action categories. 

The \textbf{Two Hands and Object (H2O)} \cite{Kwon_2021_ICCV} dataset contains 571,645 frames from 4 participants performing 37 actions in environments like a hall, office, and kitchen. It includes 42 hand pose nodes and 8 object joints, capturing actions like grabbing, placing, and pouring. The dataset is recorded at 30 FPS with synchronized RGB and depth images using multiple cameras mounted on a headset.

To validate our framework’s open-world capabilities, we evaluate across three scenarios mirroring real-world deployment: 1. \textbf{Closed-set recognition} (easy): testing exclusively on known in-distribution (ID) actions, measuring basic classification capability. 2. \textbf{Open-set recognition} (medium): testing on mixed ID and OOD actions, with OOD treated as a unified “unknown” class. 3. \textbf{OOD classification} 
(difficult): testing exclusively on unknown actions. We use an 80:20 ID:OOD split in both the Bimacs and H2O datasets, where 80\% of the data consists of known action classes (ID) for training, and the remaining 20\% consists of unseen action classes (OOD) for testing. In Bimacs, classes 11–13 are designated as OOD, whereas in H2O, classes 30–37 represent OOD actions.

The openness \(O\) of the task is calculated using the formula:
\(
    O = 1 - \sqrt{2 \cdot N_{\text{train}}/({N_{\text{test}} + N_{\text{target}}})}
\), yielding a value of approximately 6.3\%. While lower than typical open-set recognition (OSR) tasks, which focus on distinguishing known from unknown samples, it is well-suited for our open-world recognition (OWR) problem. OWR not only detects unknown samples but also differentiates among them. 

\subsection{Experimental settings}

\textbf{Quantitative Analysis:} We evaluate our model using a range of metrics that assess both closed-set and open-set performance. We use the Top 1 accuracy (ACC$_{close}$) for closed-set tasks to measure classification performance on known samples. In the open-set scenario, we extend this to include unknown samples, computing open-set Top 1 accuracy (ACC$_{open}$), and the F1@K score for temporal action segmentation. 

Additionally, we assess Out-of-Distribution detection using Area Under the Receiver Operating Characteristic Curve (AUROC) and quantify the model’s ability to distinguish between ID and OOD samples by measuring separability across all classification thresholds. A higher AUROC indicates better OOD detection.  We evaluate classification accuracy ($ACC_{OOD}$) exclusively on samples identified as OOD using their ground-truth labels, ensuring the metric directly reflects the model’s ability to classify known OOD instances. While AUROC and $ACC_{OOD}$ individually evaluate detection and classification, neither captures the model’s combined ability to first detect and then classify OOD samples. We thus combine AUROC and OOD accuracy into the harmonic mean score, computed as: \(h_{score} = {2}/({\{AUROC}^{-1}+{ACC_{OOD}^{-1}})\), penalizing imbalanced performance to ensure robust real-world OOD handling. 

\textbf{Qualitative Analysis:} To complement the quantitative evaluation, we use t-SNE to visualize feature embeddings from the Temporal Pyramid Pooling (TPP) layer, offering insights into how well the model separates In-Distribution and Out-of-Distribution samples. The ideal outcome is clear clustering of ID samples and distinct OOD categories. This helps assess the model's generalization and ability to distinguish unknown actions.

Experiments are conducted using PyTorch on an NVIDIA RTX 2070 GPU. We use stochastic gradient descent (SGD) with Nesterov momentum (0.9) and the loss formulated in equation (10). The batch size is 16, with a weight decay of 0.001. Training spans 60 epochs, and the model with the best validation accuracy and lowest loss is selected. The initial learning rate is 0.1 with exponential decay (rate 0.95). The Temporal Clustering Loss anchor magnitude is set to 20, applied to the Temporal Efficient Upsampling layer, and the Mixup \(\alpha\) is set to 0.2.

\begin{table*}[htpb]
  \centering
  \caption{Comparison of the frame-wise performance and open-set F1@K score of our method against other state-of-the-art frameworks. Bimacs rows correspond to results on the Bimanual Actions dataset~\cite{Dreher2019LearningOR}, while H2O rows correspond to the 2 Hands and Object dataset~\cite{Kwon_2021_ICCV}.}

    \begin{tabular}{l|l|l|c|c c c c | c c c}
      \toprule
      \multirow{2}{*}{Dataset} & \multirow{2}{*}{Methods} & \multirow{2}{*}{$\#$Param$^b$} & \multicolumn{1}{c|}{Closed-Set} & \multicolumn{4}{c|}{Open-Set} & \multicolumn{3}{c}{Out-of-Distribution} \\
      & & & ACC$_{close}$ & ACC\(_{open}\)& \(F1@10\) & \(F1@25\) & \(F1@50\) & \(AUROC\) & \(ACC_{OOD}\) & \(h_{score}\) \\
      \midrule
      \multirow{7}{*}{Bimacs} 
      & PGCN \cite{xing2022understanding} & 5.42 M & 79.61 & 70.39 & 83.70 & 81.88 & 73.10 & 50.00 & $\mathbf{85.86}$ & 63.21 \\
      & ST-GCN+TPP \cite{Yan2018SpatialTG} & 5.05 M & 74.16 & 65.77 & 88.62 & 86.73 & 77.35 & 50.00 & 35.26 & 41.36 \\
      & AGCN+TPP \cite{2sagcn2019cvpr} & 5.42 M & 73.93 & 65.62 & 85.63 & 83.21 & 72.06 & 51.40 & 36.65 & 42.79 \\
      & CTR-GCN+TPP \cite{9710007} & 3.36 M & 74.11 & 65.44  & 89.97 & 88.49 & 65.44& 50.00 & 44.62 & 47.16 \\ 
      &  UQ-TFGCN \cite{xing2024understanding} & 20.13 M & 83.38 & 74.98 & 90.25 & 88.28 & 78.83 & 51.79 & 69.77 & 59.25 \\
      & UQ-TFGCN+$^b$ & 20.13 M & 84.44 & 81.94 & 93.72 & 92.76 & 87.10 & 69.78 & 69.78 & 69.78 \\
      & \textbf{EPGCN} & 5.64 M & $\mathbf{85.21}$ & $\mathbf{85.71}$ & $\mathbf{95.10}$ & $\mathbf{95.08}$ & $\mathbf{90.03}$ & $\mathbf{84.62}$ & 84.69 & $\mathbf{84.65}$ \\
      \midrule
      \multirow{4}{*}{H2O} 
      & PGCN~\cite{xing2022understanding} & 5.42 M & 81.72 & 61.17 & 85.58 & 80.29 & $\mathbf{74.02}$ & 50.00 & 47.75 & 48.85 \\
      & ST-GCN+TPP~\cite{Yan2018SpatialTG} & 5.05 M & 72.99 & 54.08 & 76.24 & 70.49 & 55.28 & 50.00 & 52.37 & 51.57 \\
      & AGCN+TPP~\cite{2sagcn2019cvpr} & 5.42 M & 79.48 & 61.95 & 83.97 & 78.52 & 70.45 & 50.00 & 61.38 & 55.11 \\
      & UQ-TFGCN~\cite{xing2024understanding} & 20.13 M & $\mathbf{88.09}$ & 65.51 & $\mathbf{88.56}$ & $\mathbf{82.69}$ & 72.83 & 50.00 & 66.10 & 56.93\\
      & \textbf{EPGCN} & 5.64 M & 81.95 & $\mathbf{74.94}$ & 86.50 & 80.00 & 73.06 & $\mathbf{72.97}$ & $\mathbf{84.28}$ & $\mathbf{78.22}$ \\
      \bottomrule
      
    \end{tabular}
  
  \label{table:sota}
    \begin{tablenotes}
        \item[a]$^a$ The best results of each dataset are in \textbf{bold}. For the Bimacs dataset, the models are evaluated on the subject 1 testset.
        \item[b]$^b$ The number of learnable parameters.
        \item[c]$^c$ The original UQ-TFGCN model is trained with the mixup data augmentation and the \(\mathcal{L}_{TC}\) loss.
  \end{tablenotes}
  \vspace*{-1.5\baselineskip}

\end{table*}

\subsection{Ablation studies}
We systematically evaluate the impact of design choices on open-world performance using the Bimanual Actions dataset~\cite{Dreher2019LearningOR}. Starting with the Pyramid Graph Convolutional Network (PGCN) baseline~\cite{xing2022understanding}, we incrementally introduce components to isolate their contributions.

As shown in Table~\ref{table:bimacs}, we investigate the impact of various design choices on our proposed framework by starting with the baseline PGCN model and progressively introducing modifications. First, we examine the impact of Mixup and Temporal Clustering Loss (\(\mathcal{L}_{TC}\)) both individually and in combination with the PGCN and TEU-based models. While adding \(\mathcal{L}_{TC}\) enhances AUROC and open-set F1@K performance, it reduces Out-of-Distribution (OOD) accuracy (\(ACC_{\text{OOD}}\)) and the \( h_{\text{score}} \). This decline occurs because \(\mathcal{L}_{TC}\) encourages clustering of unknown samples, making them less distinguishable by class. Consequently, this hinders the effectiveness of the K-means algorithm in separating different clusters. On the other hand, training with Mixup alone leads to only marginal improvements over the base models. However, the TEU-based model benefits more from Mixup compared to PGCN due to its attention-based upsampling mechanism which better preserve both discriminative features and fine-grained motion. Notably, the combination of Mixup and \(\mathcal{L}_{TC}\) produces the best results for both PGCN and TEU-based models but TEU-based model performs slightly better in the more difficult categories such as F1@50 and \( h_{\text{score}} \). This is because Mixup mitigates the issue faced by adding \(\mathcal{L}_{TC}\) where unknown samples form tight clusters in the feature space. By injecting uncertainty, Mixup disrupts this clustering effect by acting as a regularizer that improves generalization to unknown samples.  

Next, we examine the impact of applying $\mathcal{L}_{TC}$ at different encoder layers. Specifically, we apply $\mathcal{L}_{TC}$ at the 4th encoder layer alongside Mixup to assess its effect on clustering features at an earlier stage of the network.  This is to determine whether enforcing temporal clustering constraints during intermediate feature extraction, rather than at the feature fusion stage could enhance robustness. However, this configuration does not yield significant improvements over our final framework. 


As shown in Table~\ref{table:bimacs}, our EPGCN framework, which integrates Mixup, $\mathcal{L}_{TC}$, and a TEU module for attention-based upsampling, achieves superior performance compared to the baseline PGCN on the Bimacs dataset. Specifically, EPGCN consistently outperforms PGCN across all open-set F1@K metrics, with a notable 16.9\% improvement in F1@50 and a 21.4\% increase in the h-score. Alternative modifications fail to yield substantial gains, reinforcing that the combination of Mixup, $\mathcal{L}_{TC}$, and TEU-based upsampling in EPGCN is optimal for both In-Distribution (ID) and Out-of-Distribution (OOD) tasks.


\begin{figure*}[!t]
  \centering
  \subfloat[PGCN]
  {\includegraphics[width=0.44\textwidth]{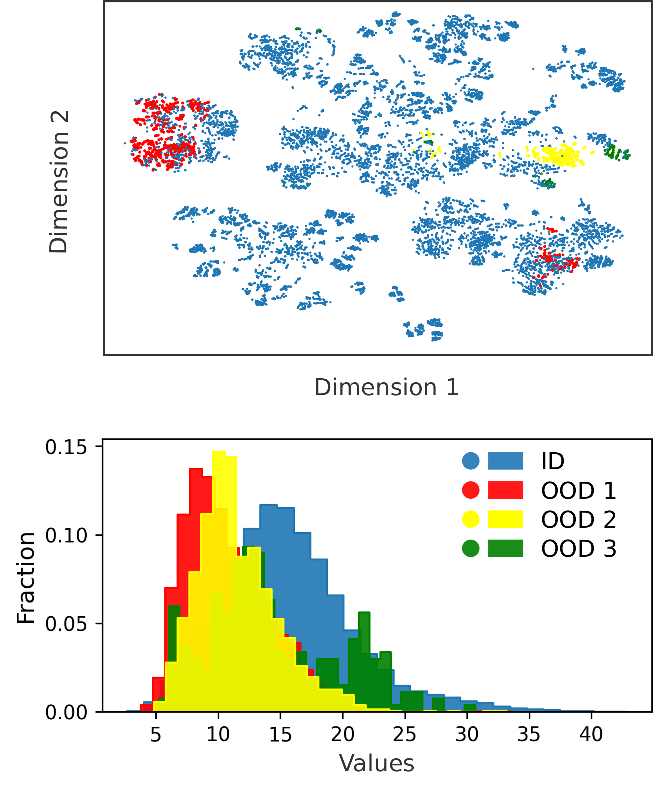}\label{fig:bimacstsne1}}
  \subfloat[EPGCN]{\includegraphics[width=0.44\textwidth]{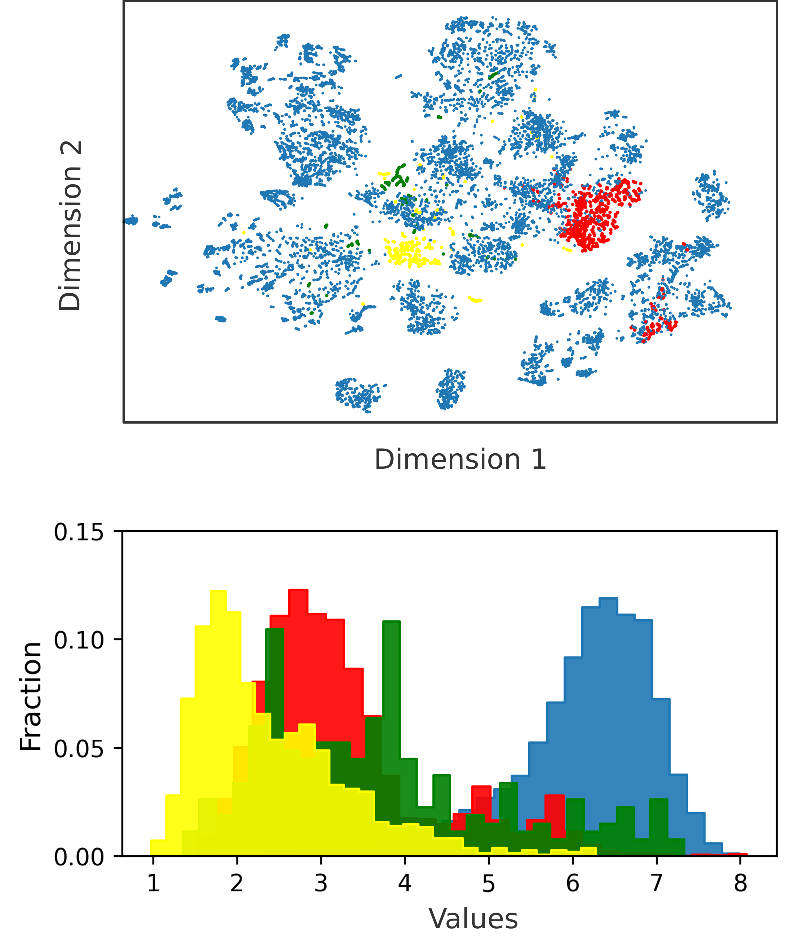}\label{fig:bimacstsne2}}
  \caption{The top row presents t-SNE visualizations of TPP feature distributions, where Dimension 1 and 2 correspond to the 2 downsampled feature dimensions. The bottom row illustrates the feature map values distributions of all In-Distribution (ID) samples against individual Out-of-Distribution (OOD) classes for both the PGCN baseline and the EPGCN framework on the Bimacs dataset. The box represents the boundary of the embedding distribution for the respective OOD class, indicating the region where its samples are located.}
  \label{fig:bimacs_tsne}
    \vspace*{-1.5\baselineskip}
\end{figure*}

\subsection{Comparison with the state-of-the-art}

The proposed EPGCN framework is compared with state-of-the-art action segmentation framework on the Bimacs \cite{Dreher2019LearningOR} and H2O \cite{Kwon_2021_ICCV} datasets. Several popular graph convolutional networks: ST-GCN \cite{Yan2018SpatialTG}, AGCN \cite{2sagcn2019cvpr}, CTR-GCN \cite{9710007}, and UQ-TFGCN \cite{xing2024understanding}. ST-GCN, AGCN and CTR-GCN are combined with the temporal pyramid pooling (TPP) decoder module since they are not originally designed for fine-grained segmentation task.

Table~\ref{table:sota} presents the results on all three tasks: close-set recognition, open-set segmentation, and Out-of-Distribution classification, where the top and bottom halves correspond to the performance on the Bimacs and H2O dataset, respectively. Its seen that our EPGCN framework outperforms all other frameworks on both datasets.  UQ-TFGCN, which incorporates spectral normalization in its residual layers to preserve feature-space distances for OOD detection, ranks second. Notably, EPGCN improves upon UQ-TFGCN in F1@50 and \( h_{\text{score}} \) by 11.2\% and 25.4\%, respectively. This significant improvement stems from UQ-TFGCN's focus on covariate shift, as it was evaluated on a noisy Bimacs dataset and the IKEA Assembly dataset \cite{BenShabat2020TheIA}, which is semantically dissimilar to Bimacs due to its 2D spatial representation rather than 3D. However, when UQ-TFGCN is trained with Mixup and our proposed Temporal Clustering Loss, F1@50 and \( h_{\text{score}} \) increase by 8.3\% and 10.5\% compared to training with UQ-TFGCN alone. This confirms the effectiveness of our proposed components in improving generalization in open-world scenarios.

The effectiveness of EPGCN is further validated on the H2O dataset, where it demonstrates superior performance across all OOD classification metrics, particularly in \( h_{\text{score}} \). Our framework exhibits a slight decline in F1@25 and F1@50 compared to the baseline PGCN model. Moreover, UQ-TFGCN attains higher F1@10 and F1@25 scores due to its higher closed-set accuracy by benefiting from a relatively easier overlapping ratio. Nevertheless, the substantial gain in \( h_{\text{score}} \)  underscores the advantage of EPGCN. Since \( h_{\text{score}} \) is a crucial metric for open-world action segmentation, as it evaluates a model’s ability to distinguish OOD frames between different classes, these results highlight the contributions of EPGCN’s three key components.

\subsection{Qualitative results}

Fig.~\ref{fig:bimacs_tsne} presents the t-SNE visualizations of the extracted features from the TPP layer for the PGCN baseline and our EPGCN framework on the Bimacs dataset. The purpose of this analysis is to assess the effectiveness of our proposed framework in open-world scenarios, particularly in its ability to distinguish between In-Distribution (ID) and Out-of-Distribution (OOD) samples and to separate different OOD classes in the feature space. As previously discussed, the success of our approach relies on ensuring that each OOD class forms distinct, well-separated clusters from the ID features.

In the t-SNE visualization of the PGCN baseline, OOD features (non-blue points) exhibit significant overlap with ID features (blue points). This indicates that the PGCN model struggles to confidently distinguish OOD samples from ID samples, leading to an output confidence distribution where ID and OOD samples have similar confidence scores. This is further confirmed by the confidence distribution plot, where the distributions of OOD classes 1, 2, and 3 overlap considerably with that of the ID samples, which is undesirable.

In contrast, our EPGCN framework demonstrates a clear separation between ID and OOD features in the t-SNE plot. This is primarily attributed to the Temporal Clustering Loss, which encourages OOD features to be positioned farther from ID features in the feature space. Additionally, the confidence distribution plot reveals that ID samples exhibit significantly higher output confidence compared to most OOD samples. Furthermore, OOD features are more distinctly clustered among themselves due to the regularization effect of Mixup, which prevents all OOD features from collapsing into a single cluster.

The t-SNE and confidence distribution plots validate our quantitative results, particularly improvements in AUROC, \(ACC_{OOD}\), and \(h_{score}\), confirming the effectiveness of our framework in the OWAS problem setting.


    \section{CONCLUSIONS}
\label{sec:5}

In this work, we formalize the Open-World Action Segmentation (OWAS) problem and propose a novel framework that addresses critical limitations in existing open-world recognition. Our approach employs a distance-based classifier (K-means) to automatically cluster novel actions without manual labeling, leveraging the observation that Out-of-Distribution (OOD) samples form distinct groupings. We enhance a closed-set backbone through three key innovations: (1) A Temporal Efficient Upsampling (TEU) module that fuses multi-scale encoder features for improved temporal coherence, (2) Mixup-based uncertainty injection during training to generalize across diverse OOD scenarios, and (3) Temporal Clustering Loss to enforce separable feature representations. Evaluations on public datasets demonstrate state-of-the-art performance, validating our framework's efficacy for segmenting both known and novel actions.

While this work provides a streamlined solution for labeling novel actions using only signals from known classes, it is limited by the clustering algorithm K-means, which requires predefined number of classes. Future work will focus on continual cluster refinement for seamless novel action integration, cross-modal generalization through motion data fusion to enhance real-world robustness, risk-aware novelty detection to prevent high-stakes misclassification, and further deploy the framework in safety-critical Human-Robot Interaction scenarios.


    \section*{ACKNOWLEDGMENT}
The research has been (partially) supported by the German Research Foundation (DFG), SFB 1320 EASE, CRC, University of Bremen. The research was conducted in subproject R1: NEEM-based embodied knowledge system.

	
	
	\bibliographystyle{IEEEtran}
	\bibliography{IEEEexample}

\end{document}